\documentclass[10pt,twocolumn,letterpaper]{article}

\usepackage{cvpr} 
\usepackage{float}
\usepackage{booktabs}
\usepackage{graphicx}
\usepackage{lscape}
\usepackage[section]{placeins}
\usepackage[table,xcdraw,dvipsnames]{xcolor}
\restylefloat{table}
\usepackage[dvipsnames]{xcolor}

\usepackage[pagebackref,breaklinks,colorlinks]{hyperref}

\def\paperID{7} 
\def\confName{CVPR\xspace}
\def\confYear{2024\xspace}

\title{FewShotNeRF: Meta-Learning-based Novel view Synthesis for Rapid Scene-Specific Adaptation}

\author{Piraveen Sivakumar\thanks{Equal contribution}\\
University of Moratuwa\\
{\tt\small piraveen.18@cse.mrt.ac.lk}
\and
Paul Janson\footnotemark[1]\\
University of Moratuwa\\
{\tt\small paul.18@cse.mrt.ac.lk}
\and
Jathushan Rajasegaran\\
UC Berkeley\\
{\tt\small jathushan@berkeley.edu}
\and
Thanuja Ambegoda\\
University of Moratuwa\\
{\tt\small thanuja@cse.mrt.ac.lk}
}

\begin{document}
\maketitle
\begin{abstract}


In this paper, we address the challenge of generating novel views of real-world objects with limited multi-view images through our proposed approach, \textbf{FewShotNeRF}. Our method utilizes meta-learning to acquire an optimal initialization, facilitating rapid adaptation of a Neural Radiance Field (NeRF) to specific scenes. The focus of our meta-learning process is on capturing shared geometry and textures within a category, embedded in the weight initialization. This approach expedites the learning process of NeRFs and leverages recent advancements in positional encodings to reduce the time required for fitting a NeRF to a scene, thereby accelerating the inner loop optimization of meta-learning. Notably, our method enables meta-learning on a large number of 3D scenes to establish a robust 3D prior for various categories. Through extensive evaluations on the Common Objects in 3D open source dataset\citep{reizensteinCommonObjects3D2021}, we empirically demonstrate the efficacy and potential of meta-learning in generating high-quality novel views of objects.
\vspace{-5mm}
\end{abstract}    
\vspace{-5mm}
\section{Introduction}
\label{sec:intro}
Neural radiance fields (NeRF)~\cite{mildenhallNeRFRepresentingScenes2020} have emerged as a transformative technology in the realm of novel view synthesis~\cite{leNovelViewSynthesis2000,xieNeuralFieldsVisual2022a,scharstein2003view}, particularly in the context of posed multiview images. This advancement is attributed to the utilization of a coordinate-based representation~\cite{parkDeepSDFLearningContinuous2019,meschederOccupancyNetworksLearning2019,sitzmannSceneRepresentationNetworks2019}, wherein a three-dimensional coordinate system is efficiently mapped to its corresponding color and density~\cite{mildenhallNeRFRepresentingScenes2020}. By adopting this approach, the representation of a three-dimensional scene becomes more compact and memory-efficient~\cite{sitzmannImplicitNeuralRepresentations2020,dupontCOINCOmpressionImplicit2021,dupontCOINDataAgnostic2022}. However, it is important to acknowledge that this enhancement comes at the expense of increased computational costs~\cite{mildenhallNeRFRepresentingScenes2020,mullerInstantNeuralGraphics2022}.

\begin{figure}
    \centering
    \includegraphics[width=\columnwidth]{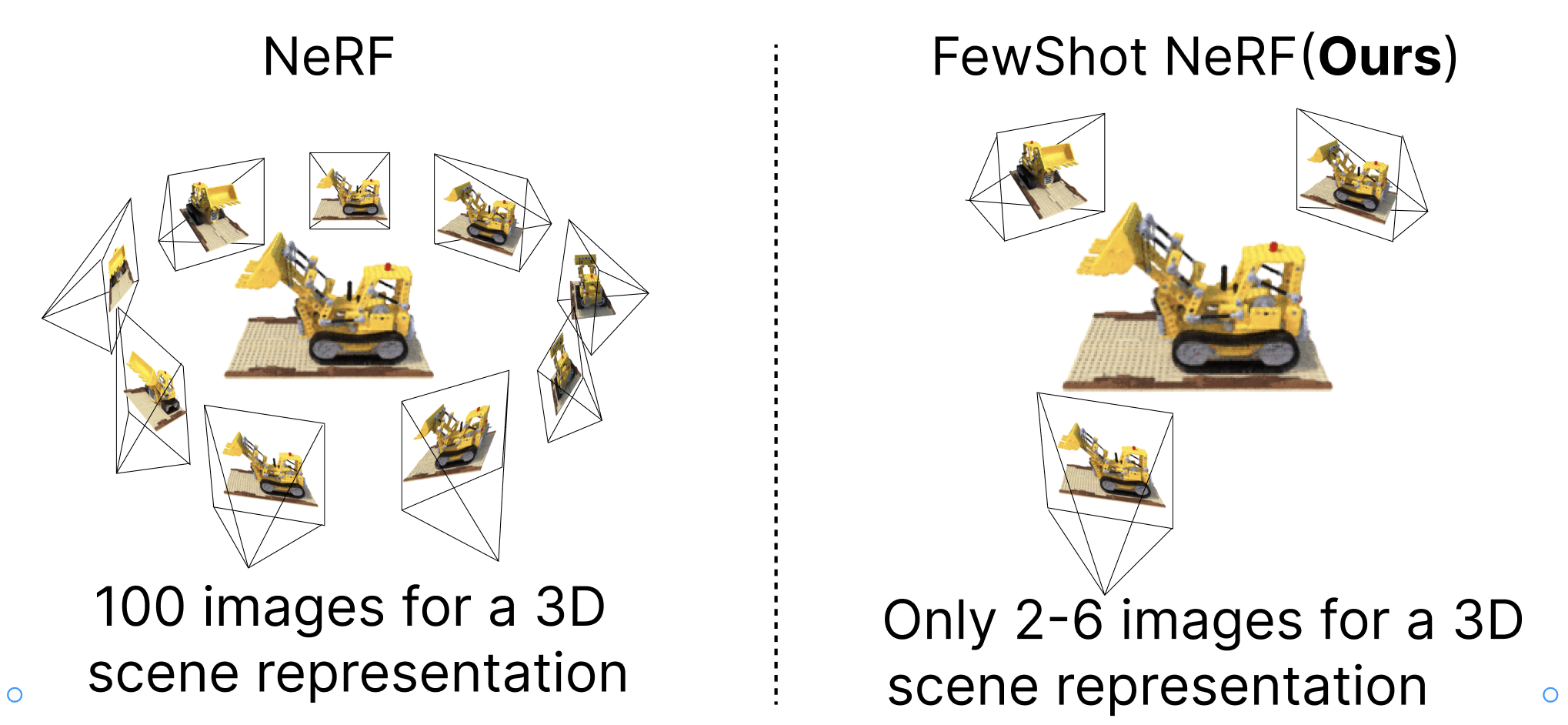}
    \caption{FewShot-NeRF: Learning Rich 3D Scenes from Minimal Camera Poses. Conventional NeRF training demands nearly 100 camera poses per scene. Our approach reduces this requirement by harnessing meta-learning to acquire an optimized initialization for NeRF. By incorporating a 3D prior into the parameter initialization, FewShot-NeRF learns a 3D scene with a minimal set of camera poses, effectively reducing frame requirements}
    \label{fig:intro}
    \vspace{-5mm}
\end{figure}

To construct a NeRF capable of generating novel views, each scene necessitates the initialization of a new model from scratch, followed by training via volume rendering~\cite{kajiyaRayTracingVolume1984} using 2D supervision provided in the form of multi-view images. Nonetheless, the efficacy of this approach is subject to certain limitations~\cite{tancikLearnedInitializationsOptimizing2021}. Firstly, it relies on large datasets containing hundreds of images capturing a single scene, which may pose challenges in terms of data acquisition and storage~\cite{xieNeuralFieldsVisual2022a}. Additionally, the computational demands associated with this methodology may be restrictive and require substantial computing resources~\cite{mullerInstantNeuralGraphics2022}. Thus, while neural radiance fields offer considerable benefits in terms of compact representation and memory efficiency, their practical implementation is hindered by the reliance on extensive datasets and the computational burdens they entail.

As researchers delved into the realm of Neural Radiance Fields (NeRFs), they recognized the importance of addressing the challenge of generalization~\cite{xieNeuralFieldsVisual2022a,sitzmannMetaSDFMetaLearningSigned2020,yuPixelNeRFNeuralRadiance2021}. Several studies have emerged, each attempting to tackle this issue from various angles. The key idea behind these attempts is to incorporate prior knowledge about the world into the initial NeRF models, enabling the learning of scene representations from a few views.

To inject these models with this prior knowledge, researchers have explored three main approaches~\cite{xieNeuralFieldsVisual2022a}. The first involves conditioning the NeRF model on a latent code~\cite{yuPixelNeRFNeuralRadiance2021,chenTransformersMetaLearnersImplicit2022}, while the second entails learning a prior initialization~\cite{tancikLearnedInitializationsOptimizing2021,sitzmannMetaSDFMetaLearningSigned2020} that facilitates rapid convergence to a scene with limited views. The third way is to use diffusion generative models to generate views and use them to train a NeRF\cite{zhouSparseFusionDistillingViewconditioned2023}. However, it is worth noting that models conditioned on a latent code may suffer from limited expressivity due to the constraints imposed by the code's size~\cite{xieNeuralFieldsVisual2022a}. That means, once developed, the restriction caused by the latent code will remain as a constraint. Diffusion-based methods rely on a 2D prior. On the other hand, gradient-based meta-learning approaches inherently maintain the full expressivity of NeRF models, thereby enabling the representation of any scene that can be captured by per-scene optimized NeRF models. The prior knowledge learned is inherently 3D.

The first work investigated the usage of gradient-based meta-learning is \cite{tancikLearnedInitializationsOptimizing2021}, specifically using Model-Agnostic Meta-Learning(MAML)~\cite{finnModelAgnosticMetaLearningFast2017} framework, to learn improved initializations for NeRFs. However, their study focused on a simplified version of NeRF that lacks view dependence, and the generalization achieved by their method was limited to three specific categories on a synthetic dataset~\cite{chang2015shapenet}. This limitation stems from the inefficient training of the vanilla NeRF and is further exacerbated by the memory requirements of the meta-learning process.

In contrast, our work aims to apply gradient-based meta-learning to NeRFs that incorporate view-dependent color output, with the objective of achieving generalization in many categories of commonly used objects. Moreover, our goal extends beyond synthetic datasets and encompasses the real-world objects shot on mobile phones~\cite{reizensteinCommonObjects3D2021}. By exploring these avenues, we seek to enhance the flexibility and adaptability of NeRF models for a wider range of scenes and categories.

 Significant modifications have been made to the architecture of NeRFs to enhance their efficiency~\cite{mullerInstantNeuralGraphics2022,martelACORNAdaptiveCoordinate2021,saragadamMINERMultiscaleImplicit2022}. One crucial aspect of NeRFs is the need for an encoding function~\cite{tancikFourierFeaturesLet2020,mildenhallNeRFRepresentingScenes2020} that maps the three-dimensional coordinate vectors to a higher-dimensional space to mitigate spectral bias\cite{rahamanSpectralBiasNeural2019}.

The original NeRF~\cite{mildenhallNeRFRepresentingScenes2020} architecture employed positional encodings inspired by Transformer~\cite{vaswaniAttentionAllYou2017} models to fulfill this requirement. However, recent studies have demonstrated that replacing these encoding functions with task-specific, learnable data structures can improve training efficiency and facilitate faster convergence. For example, \cite{liuNeuralSparseVoxel2020} and \cite{takikawaNeuralGeometricLevel2021} have presented approaches that utilize modified encoding functions, resulting in accelerated training and convergence. These modifications have been shown to be beneficial in terms of computational efficiency.

Additionally, \cite{mullerInstantNeuralGraphics2022} proposed a multi-resolution hash encoding function that drastically reduced the training time required for NeRF convergence by several orders of magnitude. This advancement, coupled with improvements in ray tracing algorithms and efficient implementation techniques, contributed to overall efficiency enhancements in NeRF models. Notably, the introduction of the hash encoding function ensured convergence with a significantly lower number of iterations, further optimizing the training process.

Motivated by the aforementioned findings in the existing literature, our paper aims to make three significant contributions:
\begin{enumerate}
    \item First, we propose the utilization of hash encoding as a way to accelerate the meta-learning process. This increases the feasibility of meta-learning on a large number of scenes.
    \item To evaluate the effectiveness of our proposed method, we conduct extensive experiments on categories of real-world objects. By employing a diverse set of object categories, we can assess the performance and generalization capabilities of our approach in a realistic and practical context.
    \item We investigate the efficacy of meta-learning in acquiring a 3D prior and explore its potential for generating novel views independently, without reliance on external 2D priors.
\end{enumerate}
\section{Related Work}
\subsection*{Neural Fields/Implicit Neural Representation}
Implicit Neural Representations are computational models that establish a mapping between input coordinates and signal values, enabling the encoding of 2D or 3D scenes within coordinate networks. These networks have found extensive applications in various visual learning tasks, including image representation \cite{chenLearningContinuousImage2021,stanleyCompositionalPatternProducing2007}, 3D scene reconstruction from 2D images \cite{parkDeepSDFLearningContinuous2019,meschederOccupancyNetworksLearning2019}, imaging inverse problems~\cite{sunCoILCoordinatebasedInternal2021}, and multi-view synthesis~\cite{mildenhallNeRFRepresentingScenes2020}.
These neural networks exhibit a bias towards low spatial frequency functions. In order to address this spectral bias inherent in neural networks,~\cite{rahamanSpectralBiasNeural2019} proposes a solution that leverages Fourier analysis to capture higher frequency functions. Another technique called positional embedding, initially employed in Natural Language Processing\cite{chowdhary2020natural}, has been adopted to map input coordinate vectors into embedded coordinated vectors. Sinusoidal embedding~\cite{vaswaniAttentionAllYou2017} and Fourier features, in conjunction with positional embedding, have been widely utilized in neural fields to capture higher frequency signals~\cite{tancikFourierFeaturesLet2020}.
~\cite{sitzmannImplicitNeuralRepresentations2020} introduces a method that replaces monotonic non-linearities with periodic nonlinearities to achieve this objective.
\subsection*{Novel View Synthesis}
Novel View Synthesis(NVS) pertains to the generation of a new viewpoint of a scene based on a given set of input camera images captured from various poses. Earlier approaches in the field, as discussed in \cite{LightFieldRendering}, were capable of producing photorealistic views; however, they heavily relied on densely captured images. Recent advancements, as highlighted in \cite{mildenhallNeRFRepresentingScenes2020} and \cite{liuNeuralSparseVoxel2020}, have made significant progress in novel view synthesis by utilizing 3D representations within neural networks, requiring fewer input images. Nevertheless, these methods necessitate multiple camera views for a single scene to fit a particular model, resulting in lengthy training times. Furthermore, a distinct model optimization process is required for each scene \cite{yuPixelNeRFNeuralRadiance2021,johariGeoNeRFGeneralizingNeRF2022}.

To address the computational cost, a recent work by Muller \etal \cite{mullerInstantNeuralGraphics2022} introduced an innovative approach capable of training a model within a few minutes. Additionally, concurrent research efforts \cite{yuPlenoxelsRadianceFields2021} have also aimed to enhance both training time and accuracy. In this study, we investigate novel view synthesis using a limited number of training samples, utilizing the approach presented in \cite{mullerInstantNeuralGraphics2022} as our base model.

\subsection*{Meta Learning}

Meta-learning~\cite{vilalta2002perspective} is a machine-learning paradigm that involves pre-training a model to acquire the ability to learn efficiently. Notably, Model Agnostic Meta-Learning (MAML)~\cite{finnModelAgnosticMetaLearningFast2017} and Reptile~\cite{nicholFirstOrderMetaLearningAlgorithms2018} are optimization-based algorithms commonly used in meta-learning. In addition to these, there exist other variants of meta-learning algorithms, such as those described in ~\cite{antoniouHowTrainYour2022, rajasegaranITAMLIncrementalTaskAgnostic2020, rajasegaran2022fully, rajasegaran2020meta}. Gradient-based meta-learning employs outer loops of Stochastic Gradient Descent (SGD) to learn an improved initialization, enabling fast convergence when faced with new instances of the same task during testing \cite{finnModelAgnosticMetaLearningFast2017}. Specifically, this approach has been applied to tasks related to neural representation, such as effectively fitting tasks to represent signed distance fields \cite{parkDeepSDFLearningContinuous2019}, with \cite{tancikLearnedInitializationsOptimizing2021a} introducing the concept of learned initialization as the first work to address gradient-based meta-learning for Neural Radiance Fields (NeRFs). However, their experiments were constrained to simplified NeRF architectures and evaluation settings.

Another approach within the realm of meta-learning involves learning a hyper network as a prior for model initialization. Hypernetworks~\cite{ha2016hypernetworks} refer to neural networks that produce weights for another neural network. Several studies have utilized hyper networks to estimate weights for implicit neural networks~\cite{chenLearningContinuousImage2021,skorokhodovAdversarialGenerationContinuous2021}. However, these early works focused solely on developing models with 2-dimensional output or models with 3D supervision.

A recent proposal \cite{chenTransformersMetaLearnersImplicit2022} suggests employing a transformer as a hyper network, drawing inspiration from the similarities between gradient-based meta-learning and the residual connections found in transformers. In our research, our objective is to apply meta-learning to learn the initialization of view-dependent NeRFs, and subsequently evaluate its performance in a challenging setting that has not been extensively explored before.

\section{Method}

Reconstructing a scene in 3D faithfully requires lots of multi-view images. Given enough multi-view images NeRF~\cite{mildenhallNeRFRepresentingScenes2020} and other multi-view reconstruction methods can reconstruct the scene with reliable 3D shapes and texture. Essentially, more views add more constraints to the optimization problem, thus creating a faithful reconstruction of the real scene. However, if we have very few images of a scene, for example, if we have only one side view of a car, then we need to rely on some additional information such as cars are usually symmetric to get some estimates of the 3D shape. Therefore, the lesser the number of views we have, we need to rely on the additional priors about the world to solve this under-constrained problem. 

In this work, we operate on a limited number of views (eg 2-6 views). This requires learning additional priors about the world, such as symmetries, smooth surfaces and even sometimes man-made priors like objects are usually rectangular, etc. For example, apples have a solid shape prior almost all of them are sphere-shaped, and plants for example share some priors on the texture, most of the leaves are usually green. This extra knowledge about the world can be enforced by allowing the model to only render a scene that lies on the manifold of real images. This explicitly adds more constraint to the texture of the underlying 3D shape thus, reducing the plausible reconstructions. The same can be applied for 3D shape priors, with points clouds. On the other hand, in this work, we propose \textbf{FewShotNeRF} to learn additional priors implicitly from the weights of the NeRF model. Our method heavily relies on the understanding of NeRF and Metalearning, and we briefly discuss these two ideas next.  

\begin{figure*}[!ht]
\centering
  \includegraphics[width=0.85\textwidth]{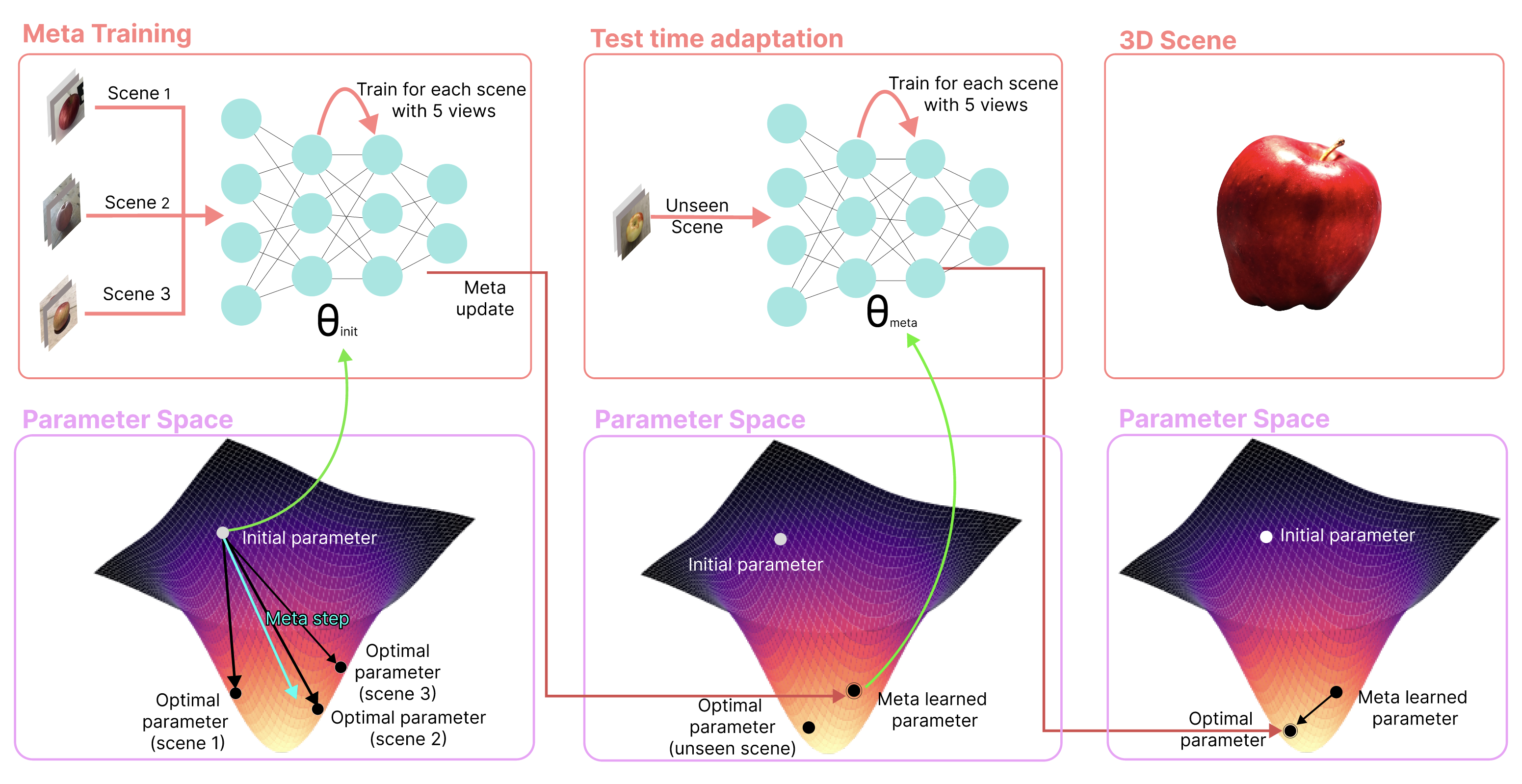}
  \caption{Method Overview: (Left) Our approach is rooted in the concept of meta-learning for initialization. We dynamically adjust the initialization by shifting it closer to the optimal parameters derived from NeRFs fitted to various scenes within the same category. This update leverages an extensive range of category-related scenes to imbue geometric resemblances into the initialization. (Center) During testing, we employ 2 to 6 images from distinct viewpoints, initiating NeRF fitting with the learned initialization. (Right) The resulting NeRF model facilitates the synthesis of novel views for the depicted scene.}
  \label{fig:main_fig}
\end{figure*}

\noindent\textbf{Neural Radiance Fields (NeRF):} NeRFs synthesis realistic 3D scenes by directly learning volumetric representations from 2D images. Unlike traditional methods such as point clouds or meshes, NeRF excels at capturing fine details and complex occlusions. NeRF employs a neural network, specifically a multi-layer perceptron (MLP), to approximate the volumetric scene representation. This network maps 3D coordinates to radiance values, yielding highly realistic scene synthesis. To render images from NeRF, it employs ray marching casting rays from a virtual camera and integrating radiance values along the ray's path. NeRF's training requires a dataset with 2D images and camera poses. Through differentiable rendering and gradient descent, it optimizes MLP parameters by minimizing differences between rendered and ground truth images.

\noindent\textbf{Meta Learning:} Its objectives are effective adaptation and generalization. Here models are trained to adapt quickly to a task such that newer tasks can be solved with limited resources. Tasks within Meta Learning are typically structured within a task distribution or "meta-dataset." Individual tasks represent distinct learning problems, while the meta-dataset spans a range of tasks, facilitating the acquisition of transferable knowledge. 


\subsection*{Meta-Learning Algorithms}

In our study, we conducted a comprehensive evaluation of three prominent meta-learning algorithms, namely Reptile, First Order MAML, and Second Order MAML. Each algorithm was assessed based on multiple criteria, including computational efficiency, memory consumption, and meta-training adaptation steps. Through this systematic evaluation, we aimed to identify the most suitable algorithm for our task.

\noindent\textbf{Reptile}: Reptile emerged as a compelling choice due to its favorable computational cost and efficient training process. It demonstrated remarkable performance in scenarios with a large number of meta-training adaptation steps (\(200\)). Additionally, Reptile exhibited fast convergence and required relatively less memory consumption, making it well-suited for our experimental setup. However, it's important to note that Reptile's performance might plateau with a limited number of adaptation steps, leading to the need for a larger step count to achieve optimal results.

\begin{equation}
\theta' = \theta + \alpha (\theta'_{\text{meta}} - \theta)
\end{equation}

Here, \(\theta\) represents the initial model parameters, \(\theta'_{\text{meta}}\) is the meta-learned model parameters, and \(\alpha\) is the meta-learning rate.

\noindent\textbf{First Order MAML}: First Order MAML exhibited competitive performance, sharing similarities with Reptile in terms of the number of meta-training steps required for adaptation. However, one notable distinction was its higher memory consumption, which could impact its scalability for larger datasets or more complex tasks. While it offered comparable results, the increased memory requirements might limit its practical utility in certain scenarios.

\begin{equation}
\theta' = \theta - \alpha \nabla_{\theta} \mathcal{L}_{\text{meta-train}}(f_{\theta})
\end{equation}

Here, \(\nabla_{\theta}\) denotes the gradient with respect to the model parameters \(\theta\), and \(\mathcal{L}_{\text{meta-train}}(f_{\theta})\) represents the meta-training loss of the model \(f_{\theta}\).

\noindent\textbf{Second Order MAML}: Second Order MAML demonstrated a unique profile, showcasing the potential for achieving comparable outcomes with a relatively low number of meta-training adaptation steps (\(10\)). However, its utility was offset by very high memory consumption, which could hinder its applicability in resource-constrained environments. Despite its capacity to achieve competitive results, the trade-off between memory usage and performance improvement should be carefully considered.

\begin{equation}
\theta' = \theta - \alpha (\nabla_{\theta} \mathcal{L}_{\text{meta-train}}(f_{\theta}) + \beta \nabla_{\theta}^2 \mathcal{L}_{\text{meta-train}}(f_{\theta}))
\end{equation}

In this equation, \(\beta\) represents the second-order meta-learning rate, and \(\nabla_{\theta}^2\) denotes the Hessian matrix (second-order gradient) with respect to the model parameters \(\theta\).

\subsection*{FewShotNeRF}

We formally define our problem as follows: we consider a set of images denoted as $\mathbf{I} = \{I_1, I_2, ..., I_n\}$ along with their corresponding poses $\mathbf{P} = \{P_1, P_2, ..., P_n\}$. Additionally, we have a fixed budget of $m$ optimization steps allocated for a specific scene $\mathcal{S}$. It is important to note that the size of $\mathbf{I}$ is limited to 2-6 images depending on the setup ($2\leq n \leq 6$). We aim to learn a function $f_\theta$ that can be utilized to generate a new image set $\mathbf{I}'$ using their corresponding poses $\mathbf{P}'$. Importantly, there should be no overlap between the poses in $\mathbf{P}$ and $\mathbf{P}'$, denoted by $\mathbf{P} \cap \mathbf{P}' = \Phi$. The function $f_\theta$ follows the same architecture as described in NeRF~\cite{mildenhallNeRFRepresentingScenes2020}, and the generated images are produced accordingly. The objective of training $f_\theta$ is to ensure that the generated image set $\mathbf{I}'$ matches the ground truth images $\mathbf{I}'_{gt}$ for the corresponding poses. Here, $\mathbf{I}'_{gt}$ represents the ground truth images for the given poses $\mathbf{P}'$. 
\vspace{-0.05cm}
From the set of scenes, first, we sample a random $k$ number of scenes $\{\mathcal{S}_1, \mathcal{S}_2, ... \mathcal{S}_k\}$. For each scene in the randomly sampled set, we learn a NeRF optimization in the inner loop.
\begin{equation}
    {\theta}^i_j = NeRF(\mathcal{S}_j, {\theta}^i)
\end{equation}
\vspace{-0.05cm}
Here, $\theta^i_j$ is the optimized parameters of the MLP, for the scene $\mathcal{S}_j$, where the optimization is initialized with the parameters ${\theta}^i$ at $i$th meta-learning iterations. This process defines the inner loop optimization of the meta-learning algorithm, and here the inner loop is equivalent to the NeRF optimization. For the outer loop optimization, we use Reptile~\cite{nicholFirstOrderMetaLearningAlgorithms2018} as our main meta-learning algorithm, and we abilate performance of FewShotNeRF when optimizing with MAML~\cite{finnModelAgnosticMetaLearningFast2017} and FOMAML~\cite{finnModelAgnosticMetaLearningFast2017}. After optimizing the task-specific parameters in the inner loop, we update the meta-parameters in the out loop using a simple weighted sum of inner loop gradients as shown in Fig~\ref{fig:main_fig}.
\begin{equation}
    {\theta}^{i+1} = {\theta}^{i} - \alpha \sum_{j=1}^k ({\theta}^{i}_j - {\theta}^{i})
\end{equation}
\vspace{-0.05cm}
Here,$i$ is the number of meta-learning iterations. $\alpha$ is the learning rate of the meta-learning algorithm. We train the meta parameters for over a few hundred iterations and then at test time given a few views of a novel scene, we apply inner loop optimization (NeRF) to the meta initialized parameters $\theta$ and then render novel views using the inner-loop optimized parameters.

\section{Experiments}


We conducted a comprehensive evaluation of our approach using the publicly available CO3D dataset \cite{reizensteinCommonObjects3D2021}, which encompasses real-world multi-view objects. To assess the effectiveness of our method in addressing the challenge of generalizing Neural Radiance Fields (NeRFs), we performed a comparative analysis against existing techniques. Our evaluation encompassed diverse scenarios characterized by different quantities of input views employed for NeRF generation, specifically utilizing 2, 3, and 6 input frames in our experimental setups.

\begin{table*}[ht!]
\centering
\resizebox{\textwidth}{!}{%
\begin{tabular}{@{}lcccccccccc@{}}
\toprule
 \rowcolor[HTML]{FFFFFF} 
 &
  \multicolumn{1}{l}{\cellcolor[HTML]{FFFFFF}Donut} &
  \multicolumn{1}{l}{\cellcolor[HTML]{FFFFFF}Apple} &
  \multicolumn{1}{l}{\cellcolor[HTML]{FFFFFF}{\color[HTML]{1F1F1F} Hydrant}} &
  \multicolumn{1}{l}{\cellcolor[HTML]{FFFFFF}{\color[HTML]{1F1F1F} Vase}} &
  \multicolumn{1}{l}{\cellcolor[HTML]{FFFFFF}{\color[HTML]{1F1F1F} Cake}} &
  \multicolumn{1}{l}{\cellcolor[HTML]{FFFFFF}Ball} &
  \multicolumn{1}{l}{\cellcolor[HTML]{FFFFFF}{\color[HTML]{1F1F1F} Bench}} &
  \multicolumn{1}{l}{\cellcolor[HTML]{FFFFFF}{\color[HTML]{1F1F1F} Suitcase}} &
  \multicolumn{1}{l}{\cellcolor[HTML]{FFFFFF}Teddybear} &
  \multicolumn{1}{l}{\cellcolor[HTML]{FFFFFF}{\color[HTML]{1F1F1F} Plant}} \\ \midrule
\rowcolor[HTML]{FFFFFF} 
 &
  \multicolumn{1}{c}{\cellcolor[HTML]{FFFFFF}PSNR $\uparrow$ } &
  \multicolumn{1}{c}{\cellcolor[HTML]{FFFFFF}PSNR $\uparrow$ } &
  \multicolumn{1}{c}{\cellcolor[HTML]{FFFFFF}PSNR $\uparrow$ } &
  \multicolumn{1}{c}{\cellcolor[HTML]{FFFFFF}PSNR $\uparrow$ } &
  \multicolumn{1}{c}{\cellcolor[HTML]{FFFFFF}PSNR $\uparrow$ } &
  \multicolumn{1}{c}{\cellcolor[HTML]{FFFFFF}PSNR $\uparrow$ } &
  \multicolumn{1}{c}{\cellcolor[HTML]{FFFFFF}PSNR $\uparrow$ } &
  \multicolumn{1}{c}{\cellcolor[HTML]{FFFFFF}PSNR $\uparrow$ } &
  \multicolumn{1}{c}{\cellcolor[HTML]{FFFFFF}PSNR $\uparrow$ } &
  \multicolumn{1}{c}{\cellcolor[HTML]{FFFFFF}PSNR $\uparrow$ } \\ \midrule
\rowcolor[HTML]{FFFFFF} 
PixelNeRF\cite{yuPixelNeRFNeuralRadiance2021} &
  20.9 &
  20.0 &
  19.0 &
  21.3 &
  18.3 &
  18.5 &
  17.7 &
  21.7 &
  18.5 &
  19.3 \\
\rowcolor[HTML]{FFFFFF} 
NeRFormer\cite{reizensteinCommonObjects3D2021} &
  20.3 &
  19.5 &
  18.2 &
  17.7 &
  16.9 &
  16.8 &
  15.9 &
  20.0 &
  15.8 &
  17.8 \\
\rowcolor[HTML]{FFFFFF} 
ViewFormer\cite{kulhanek2022viewformer} &
  19.3 &
  20.1 &
  17.5 &
  20.4 &
  17.3 &
  18.3 &
  16.4 &
  21.0 &
  15.5 &
  17.8 \\
\rowcolor[HTML]{FFFFFF} 
EFT\cite{zhouSparseFusionDistillingViewconditioned2023} &
  21.5 &
  22.0 &
  21.6 &
  21.1 &
  19.9 &
  21.4 &
  17.8 &
  23.0 &
  19.8 &
  20.4 \\
\rowcolor[HTML]{FFFFFF} 
VLDM\cite{zhouSparseFusionDistillingViewconditioned2023} &
  20.1 &
  21.3 &
  20.1 &
  20.2 &
  18.9 &
  20.3 &
  16.6 &
  21.3 &
  17.9 &
  18.9 \\
SparseFusion\cite{zhouSparseFusionDistillingViewconditioned2023} &
  \cellcolor[HTML]{FFD689}22.8 &
  \cellcolor[HTML]{FFD689}22.8 &
  \cellcolor[HTML]{FFD689}22.3 &
  \cellcolor[HTML]{FFD689}22.8 &
  \cellcolor[HTML]{FFD689}20.8 &
  \cellcolor[HTML]{FFD689}22.4 &
  \cellcolor[HTML]{FEEED1}16.7 &
  \cellcolor[HTML]{FEEED1}22.2 &
  \cellcolor[HTML]{FFD689}20.6 &
  \cellcolor[HTML]{FEEED1}20.0 \\
FewShotNeRF(25\%) &
  \cellcolor[HTML]{FFBE42}\textbf{23.9} &
  \cellcolor[HTML]{FFBE42}\textbf{23.2} &
  \cellcolor[HTML]{FFBE42}\textbf{22.6} &
  \cellcolor[HTML]{FFBE42}\textbf{24.2} &
  \cellcolor[HTML]{FFBE42}\textbf{22.4} &
  \cellcolor[HTML]{FFBE42}\textbf{22.5} &
  \cellcolor[HTML]{FFBE42}\textbf{19.9} &
  \cellcolor[HTML]{FFBE42}\textbf{24.7} &
  \cellcolor[HTML]{FFBE42}\textbf{20.9} &
  \cellcolor[HTML]{FFBE42}\textbf{21.3} \\
FewShotNeRF(50\%) &
  \cellcolor[HTML]{FEEED1}22.6 &
  \cellcolor[HTML]{FEEED1}22.2 &
  \cellcolor[HTML]{FEEED1}21.7 &
  \cellcolor[HTML]{FEEED1}22.3 &
  \cellcolor[HTML]{FFD689}20.8 &
  \cellcolor[HTML]{FEEED1}21.0 &
  \cellcolor[HTML]{FFD689}18.5 &
  \cellcolor[HTML]{FFD689}23.0 &
  \cellcolor[HTML]{FEEED1}19.4 &
  \cellcolor[HTML]{FFD689}20.5 \\
\rowcolor[HTML]{FFFFFF} 
FewShotNeRF(75\%) &
  21.7 &
  21.3 &
  21.0 &
  20.8 &
  19.6 &
  19.9 &
  17.4 &
  21.7 &
  18.3 &
  19.3 \\
\rowcolor[HTML]{FFFFFF} 
FewShotNeRF(100\%) &
  20.5 &
  20.1 &
  20.2 &
  19.2 &
  18.2 &
  18.8 &
  16.3 &
  20.3 &
  17.0 &
  18.1
\end{tabular}
}
\caption{Results on the CO3D dataset comparing our method with baselines on categories Donut, Apple, Hydrant, Vase, Cake, Ball, Bench, Suitcase, Teddybear, and Plant. Our method outperforms most of the methods and seems to show competitive performance to the SparseFusion, while not relying on any external models. To provide a nuanced understanding of our method's performance, we went beyond conventional averaging techniques. Unlike SparseFusion, which averages results from randomly selecting 10 scenes, we conducted a thorough evaluation using 150 scenes. We then calculated averages for each quartile, breaking down our method's performance at 25\%, 50\%, 75\%, and 100\% of the scenes. For the 25\% quartile, we sorted the 150 scenes based on PSNR and selected the top 25\%, demonstrating the robustness of our method even when considering only the scenes with the highest quality. Moving to the 50\% quartile, we continued this process, ensuring a balanced representation of the dataset. At the 75\% quartile, our evaluation included scenes that ranked within the top three-quarters based on PSNR, providing a broader perspective on our method's effectiveness. Finally, the 100\% quartile encompassed the entire dataset, offering a comprehensive overview of our method's performance across the entirety of the tested scenes.}
\label{tab:co3d2view}
\end{table*}

\begin{table*}[ht!]
\centering
\begin{tabular}{@{}l|ll|ll|ll@{}}
\toprule
{\color[HTML]{000000} } &
  \multicolumn{2}{c|}{{\color[HTML]{000000} \textbf{2 Views}}} &
  \multicolumn{2}{c|}{{\color[HTML]{000000} \textbf{3 Views}}} &
  \multicolumn{2}{c}{{\color[HTML]{000000} \textbf{6 Views}}} \\ \midrule
{\color[HTML]{000000} } &
  {\color[HTML]{000000} PSNR ↑} &
  {\color[HTML]{000000} SSIM ↓} &
  {\color[HTML]{000000} PSNR ↑} &
  {\color[HTML]{000000} SSIM ↓} &
  {\color[HTML]{000000} PSNR ↑} &
  {\color[HTML]{000000} SSIM ↓} \\ \midrule
{\color[HTML]{000000} PixelNeRF\cite{yuPixelNeRFNeuralRadiance2021}} &
  {\color[HTML]{000000} 19.52} &
  {\color[HTML]{000000} 0.667} &
  {\color[HTML]{000000} 20.67} &
  {\color[HTML]{000000} 0.712} &
  {\color[HTML]{000000} 22.47} &
  {\color[HTML]{000000} 0.776} \\
{\color[HTML]{000000} NerFormer\cite{reizensteinCommonObjects3D2021}} &
  {\color[HTML]{000000} 17.88} &
  {\color[HTML]{000000} 0.598} &
  {\color[HTML]{000000} 18.54} &
  {\color[HTML]{000000} 0.618} &
  {\color[HTML]{000000} 19.99} &
  {\color[HTML]{000000} 0.661} \\
{\color[HTML]{000000} ViewFormer\cite{kulhanek2022viewformer}} &
  {\color[HTML]{000000} 18.37} &
  {\color[HTML]{000000} 0.697} &
  {\color[HTML]{000000} 18.91} &
  {\color[HTML]{000000} 0.704} &
  {\color[HTML]{000000} 19.72} &
  {\color[HTML]{000000} 0.717} \\
{\color[HTML]{000000} EFT\cite{zhouSparseFusionDistillingViewconditioned2023}} &
  \cellcolor[HTML]{FEEED1}{\color[HTML]{000000} 20.85} &
  {\color[HTML]{000000} 0.680} &
  \cellcolor[HTML]{FFD689}{\color[HTML]{000000} 22.71} &
  \cellcolor[HTML]{FEEED1}{\color[HTML]{000000} 0.747} &
  \cellcolor[HTML]{FFD689}{\color[HTML]{000000} 24.57} &
  \cellcolor[HTML]{FFBE42}{\color[HTML]{000000} 0.804} \\
{\color[HTML]{000000} VLDM \cite{zhouSparseFusionDistillingViewconditioned2023}} &
  {\color[HTML]{000000} 19.55} &
  \cellcolor[HTML]{FEEED1}{\color[HTML]{000000} 0.711} &
  {\color[HTML]{000000} 20.85} &
  {\color[HTML]{000000} 0.737} &
  {\color[HTML]{000000} 22.35} &
  {\color[HTML]{000000} 0.768} \\
{\color[HTML]{000000} SparseFusion\cite{zhouSparseFusionDistillingViewconditioned2023}} &
  \cellcolor[HTML]{FFD689}{\color[HTML]{000000} 21.34} &
  \cellcolor[HTML]{FFD689}{\color[HTML]{000000} 0.752} &
  \cellcolor[HTML]{FEEED1}{\color[HTML]{000000} 20.85} &
  \cellcolor[HTML]{FFD689}{\color[HTML]{000000} 0.766} &
  \cellcolor[HTML]{FEEED1}{\color[HTML]{000000} 23.74} &
  \cellcolor[HTML]{FEEED1}{\color[HTML]{000000} 0.791} \\
{\color[HTML]{000000} FewShotNeRF(25\%)} &
  \cellcolor[HTML]{FFBE42}{\color[HTML]{000000} \textbf{22.50}} &
  \cellcolor[HTML]{FFBE42}{\color[HTML]{000000} \textbf{0.781}} &
  \cellcolor[HTML]{FFBE42}{\color[HTML]{000000} \textbf{23.01}} &
  \cellcolor[HTML]{FFBE42}{\color[HTML]{000000} \textbf{0.781}} &
  \cellcolor[HTML]{FFBE42}{\color[HTML]{000000} \textbf{25.76}} &
  \cellcolor[HTML]{FFD689}{\color[HTML]{000000} 0.792}
\end{tabular}


\caption{Results on the CO3D dataset with 2,3,6 views on average across all of the selected 10 categories. We compare our method \textbf{FewShotNeRF} with PixelNeRF, NerFormer, SparseFusion, etc. Our experiments show that FewShotNeRF outperforms most of the comparisons and performs on par with SpareFusion. Note that the evaluation protocols are slightly different and our evaluation is more robust and stronger than a random sampling of 10 scenes.
\vspace{-7mm}
}
\label{tab:3view}
\end{table*}
\subsection{Setup}

\noindent\textbf{Dataset:} We subjected our method to evaluation using the CO3D dataset~\cite{reizensteinCommonObjects3D2021}. It has different scenes belonging to 50 categories of commonly used objects. The frames are taken from mobile phone videos. We select only 10 core categories following \cite{zhouSparseFusionDistillingViewconditioned2023} to perform our experiments. This dataset provides essential components, including relative camera poses for each frame and masks that delineate the object of interest from the background. Our selection of this dataset was motivated by the aim to investigate the efficacy of meta-learning in accelerating the learning process of Neural Radiance Fields (NeRF) within real-world scenes. This contrasts with the approach taken in \cite{tancikLearnedInitializationsOptimizing2021}, which concentrated on synthetic scenes characterized by a simplified NeRF architecture.

\noindent\textbf{Baselines:} We conducted thorough comparisons between our method and several existing approaches, all of which have been suitably adapted to accommodate the CO3D dataset as presented in \cite{zhouSparseFusionDistillingViewconditioned2023}.

Given the category-specific nature of our method, we conducted comparisons against a tailored category-specific version of Pixel-NerF~\cite{yuPixelNeRFNeuralRadiance2021}. This variant leverages pixel-wise image feature re-projection of CNN features to achieve its results. In addition, we evaluated our method against NerFormer~\cite{reizensteinCommonObjects3D2021} which is based on the feature-reprojection technique and ViewFormer~\cite{kulhanek2022viewformer} which is based on autoregressive generation.

Furthermore, we assessed the performance of our method against the state-of-the-art approach, Sparsefusion~\cite{zhouSparseFusionDistillingViewconditioned2023}. This method employs a diffusion-based prior to address data scarcity issues effectively. Through these comparisons, we demonstrated the distinct strengths and capabilities of our approach within the context of the CO3D dataset.



\noindent\textbf{Implementation Details:} We used a PyTorch implementation of Instant NGP~\citep{mullerInstantNeuralGraphics2022} as our backbone model. We rendered the images at 128 x 128 following \citet{zhouSparseFusionDistillingViewconditioned2023} to ease the memory constraints.
For meta-learning we adopted the Reptile~\citep{nicholFirstOrderMetaLearningAlgorithms2018} algorithm. Our method consisted of two main phases. The first phase is the Meta-learning phase and the second phase is the Test time adaptation phase. During the Meta-learning phase, We used a fixed budget of 200 inner optimization steps to adapt the model to a specific scene. We randomly sampled 5 scenes from the scenes selected for meta-learning after leaving 150 scenes for the testing phase and adapted them. The outer loop ran through 8 steps. During the Test time adaptation phase, we tested the method on the 150 scenes left out for testing, and using 2,3, or 6 frames, we adapted the model for 400 inner optimization steps. We evaluated the models using the PSNR and SSIM metrics on the remaining test frames.

\noindent\textbf{Meta-Learning Algorithm:} We opted for the Reptile meta-learning algorithm due to its suitability for our context. Learning a NeRF involves substantial computational demands, and incorporating memory-intensive algorithms like the one proposed in \cite{finnModelAgnosticMetaLearningFast2017} would prove impractical, especially when dealing with a large dataset. To further understand that, We systematically evaluated three distinct meta-learning algorithms, ultimately selecting Reptile based on its favorable computational cost.

\begin{table}[]
\resizebox{\columnwidth}{!}{%
\begin{tabular}{@{}lcccc@{}}
\toprule
\begin{tabular}[c]{@{}l@{}}Meta Learning \\ Algorithms\end{tabular} &
  \begin{tabular}[c]{@{}c@{}}Mean\\ (PSNR)\end{tabular} &
  \begin{tabular}[c]{@{}c@{}}Standard \\ Deviation\end{tabular} &
  \multicolumn{1}{l}{Variance} &
  \begin{tabular}[c]{@{}c@{}}No of Meta \\ Training iterations\end{tabular} \\ \midrule
Reptile \cite{nicholFirstOrderMetaLearningAlgorithms2018}          & 22.04 & 2.76 & 7.62  & 200 \\
MAML First Order\cite{finnModelAgnosticMetaLearningFast2017}  & 18.42 & 3.23 & 10.46 & 200 \\
MAML Second Order\cite{finnModelAgnosticMetaLearningFast2017} & 18.45 & 3.22 & 10.35 & 10  \\ \bottomrule
\end{tabular}%
}
\caption{This table compares the performance of three meta-learning algorithms in the Apple category. Reptile outperformed MAML First Order with the same meta-training iterations. However, MAML Second Order achieved competitive results with significantly fewer iterations, showcasing its efficiency. The findings highlight MAML Second Order's ability to achieve comparable performance with a reduced number of meta-training steps, challenging the conventional approaches of Reptile and MAML First Order.
\vspace{-10mm}
}
\label{tab:abc}
\end{table}

\noindent\textbf{Metrics:} We conduct a comparative analysis between our method and related approaches, employing quantitative metrics including Peak Signal-to-Noise Ratio (PSNR) and Structural Similarity Index (SSIM). PSNR quantifies the absolute likeness between the reconstructed view and ground truth, while SSIM evaluates the structural similarity between these views.
\subsection{Results on CO3D}
This section showcases the quantitative outcomes of our approach under diverse input view scenarios. We meticulously choose these views to align evenly along the circular trajectory followed during object capture. This selection is crucial as our method relies only on the input view signal and the initialization prior, so selecting the input views evenly spaced is pivotal in generating a coherent and realistic object representation.

\noindent\textbf{2-view Setup: } Table \ref{tab:co3d2view} shows the comparison of our method with the baselines in the challenging scenario of just 2 input views. PSNR values on each of the 10 selected categories are shown. We report the values taken from \cite{zhouSparseFusionDistillingViewconditioned2023}. \citet{zhouSparseFusionDistillingViewconditioned2023} selected only 10 scenes from the selected categories to test and report the values. Since the scene ids are not provided and it is not clear how these scenes are selected, we report the average PSNR value of all the scenes (150 scenes for each category in the test set) from the test split, for our method. Additionally, we also provide different quartile values of our method to emphasize the variation in the results. Our method performs better than NeRFormer\cite{reizensteinCommonObjects3D2021}, \cite{yuPixelNeRFNeuralRadiance2021} on various catagories. We can see competitive results to \cite{zhouSparseFusionDistillingViewconditioned2023} even without using an external prior such as a Diffusion model\cite{ho2020denoising} and our numbers are computed over 150 scenes not random 10 scenes.

\noindent\textbf{3-view Setup: } Our 3-view setup is very similar to the 2-view setting, except the only change is the model sees 3 views during the training and adoption.
%
%
Compared to the 2-view setting, the 3-view setting performs better. While this is an obvious observation in the NeRF land, this performance gain in the meta-learning setting suggests that the meta objective is able to capture strong inner-loop priors to the outer-loop. 
Detailed results, representing the average outcomes across the ten designated categories, are presented in \ref{tab:3view}.

\noindent\textbf{6-view Setup: } Extending our experimentation to a 6-view setup yields further improvements in the results. This progression is documented in Table \ref{tab:3view}. 

\subsection{Evolution of Image Quality via Meta-Training Iterations}
In our experiment, we checked how image quality improves over several iterations of meta-training. We looked at the Peak Signal-to-Noise Ratio (PSNR) for 10 different scenes in each category during each round. The results showed a clear and consistent trend. Figure \ref{fig:intro}  displayed an increase in PSNR values with more meta-training iterations. This indicates a continual improvement in image quality. Our approach, which involved evaluating multiple scenes and averaging their PSNR values, highlights the reliability of our findings and the effectiveness of iterative meta-training in making images better.
\vspace{-5mm}
\begin{figure}[h!]

    \centering
    \includegraphics[width=\columnwidth]{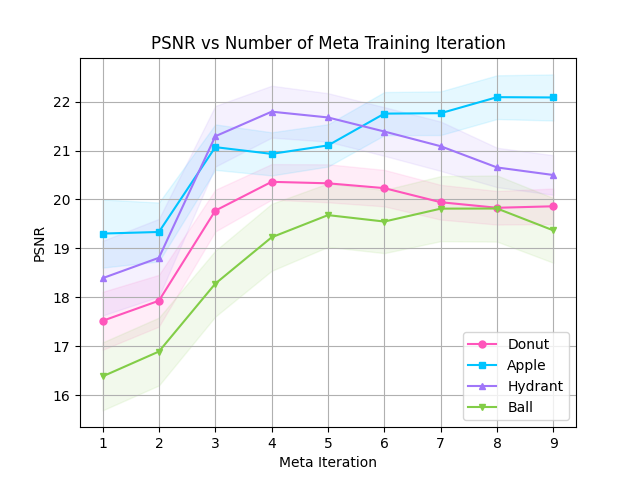}
    \caption{Evolution of PSNR Across Meta-Training Iterations. This graph illustrates the progressive increase in Peak Signal-to-Noise Ratio (PSNR) values with the number of meta-training iterations. The study includes an average of 10 scenes per category, highlighting the consistent improvement in image quality achieved through the iterative meta-training process.}
    \label{fig:intro}
    \vspace{-6mm}
\end{figure}

\section{Discussion}

In this paper, we introduce \textbf{FewShotNeRF}, an approach to generalize Neural Radiance Fields (NeRFs) for view synthesis with few input views. We make three significant contributions through this work. First, we propose to utilize hash encoding to accelerate the training of NeRF models in the inner loop of meta-learning. Second, we conduct extensive experiments on real-world object categories to evaluate the effectiveness of this method and scale the meta-training to over 300 scenes to distill the 3D priors into a single model. These experiments provide valuable insights into the feasibility and potential benefits of using hash encoding for meta-learning NeRF models. Finally, our proposed method relies on learning 3D priors only from the meta objective without relying on external models.

In conclusion, the presented findings have the potential to shed light on the NeRF generalization. The utilization of hash encoding for meta-learning initialization, along with the extensive experimental evaluations, contributes to the advancement of generalizable NeRFs. Future work could further refine and extend the proposed methodology by exploring additional enhancements and could also focus on applying FewShotNeRF to more complex and challenging scenarios like dynamic scenes.

\begin{figure*}[!ht]
\centering
   \includegraphics[width=0.95\textwidth]{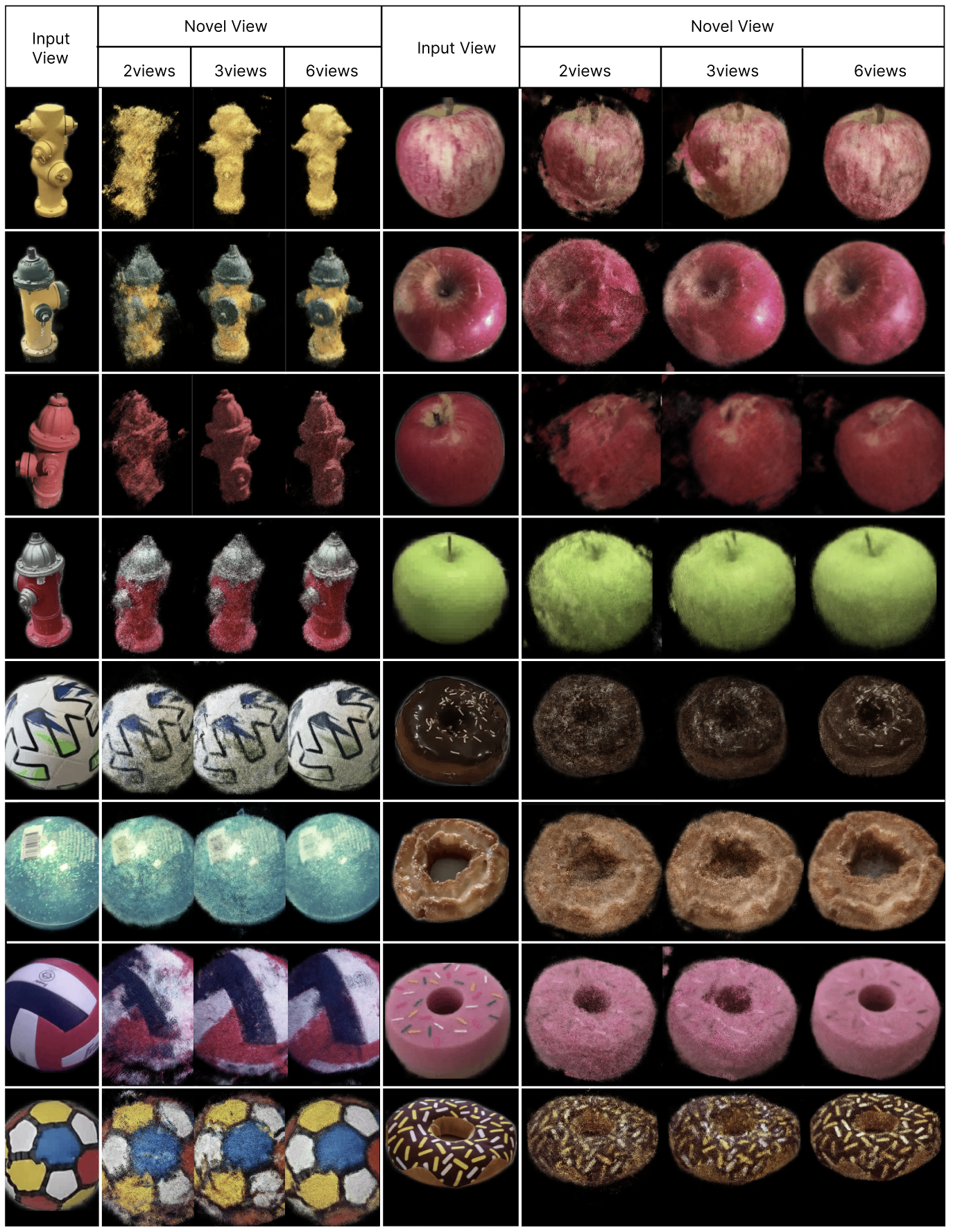}
  \caption{This sequence of images illustrates the qualitative progress achieved across four categories - hydrant, apple, ball, and donut. Beginning with the training input, followed by novel views generated using only 2, 3, and 6 training images, we witness the model's ability to enhance realism and accuracy in novel view generation.}
  \label{fig:qualitative}
\end{figure*}

\clearpage
{
    \small
    \bibliographystyle{ieeenat_fullname}
    \bibliography{main}
}
\clearpage
\setcounter{page}{1}
\maketitlesupplementary

\subsection{Metrics}
We compare our method with the baselines using the standard metrics used for Image-based comparisons. We selected PSNR as a pixel-wise comparison method and SSIM as a perceptual comparison metric. Those results are calculated as follows.
\begin{equation}
    {PSNR} = -10 \log_{10}(MSE)
\end{equation}
Where ${MSE}$ is calculated as follows
\begin{equation}
    {MSE} = \frac{1}{w \cdot h} \sum_{i=1}^{w} \sum_{j=1}^{h} (I_{\text{original}}(i, j) - I_{\text{reconstructed}}(i, j))^2
\end{equation}

Here $w$ represents the width of the image and $h$ represents the height of the image.

The equation we used to calculate the $SSIM$ is as follows.
\begin{equation}
    {SSIM(x, y)}= \frac{{(2\mu_{x}\mu_{y} + c_1)(2\sigma_{x,y} + c_2)}}{{(\mu_{x}^2 + \mu_{y}^2 + c_1)(\sigma_{x}^2 + \sigma_{y}^2 + c_2)}}
\end{equation}

Let $x$ and $y$ denote the original and reconstructed images, respectively. Furthermore, let $\mu$ and $\sigma$ represent the mean and standard deviation of pixel intensities in the images. The covariances of pixel intensities in the original and reconstructed images are denoted by $\sigma_{x, y}$. To prevent zero-division errors, constants $c_1$ and $c_2$ are introduced.

\subsection{Impact of View Count on Image Enhancement Quality}

This table \ref{tab:my-table} underscores the efficacy of meta-learning in the context of Neural Radiance Fields (NeRF), specifically in rapidly acquiring a scene's understanding with a few images while achieving high Peak Signal-to-Noise Ratio (PSNR). The comparison between meta-training with 3 views and 6 views reveals the model's capacity to excel in scene comprehension without reliance on external models. The reported PSNR values showcase the efficiency of meta-learning in enhancing image quality, particularly noteworthy for its ability to achieve substantial improvements even with a limited number of input images.

\begin{table*}[ht]

\centering

\resizebox{\textwidth}{!}{%
\begin{tabular}{@{}lrrrrrrrrrrrrrrrr@{}}
\toprule
          & \multicolumn{8}{c}{3 Views}                                   & \multicolumn{8}{c}{6 Views}                                   \\ \midrule
 &
  \multicolumn{2}{c}{Top 25\%} &
  \multicolumn{2}{c}{Top 50\%} &
  \multicolumn{2}{c}{Top 75\%} &
  \multicolumn{2}{c}{Top 100\%} &
  \multicolumn{2}{c}{Top 25\%} &
  \multicolumn{2}{c}{Top 50\%} &
  \multicolumn{2}{c}{Top 75\%} &
  \multicolumn{2}{c}{Top 100\%} \\ \midrule
 &
  \multicolumn{1}{c}{PSNR} &
  \multicolumn{1}{c}{SSIM} &
  \multicolumn{1}{c}{PSNR} &
  \multicolumn{1}{c}{SSIM} &
  \multicolumn{1}{c}{PSNR} &
  \multicolumn{1}{c}{SSIM} &
  \multicolumn{1}{c}{PSNR} &
  \multicolumn{1}{c}{SSIM} &
  \multicolumn{1}{c}{PSNR} &
  \multicolumn{1}{c}{SSIM} &
  \multicolumn{1}{c}{PSNR} &
  \multicolumn{1}{c}{SSIM} &
  \multicolumn{1}{c}{PSNR} &
  \multicolumn{1}{c}{SSIM} &
  \multicolumn{1}{c}{PSNR} &
  \multicolumn{1}{c}{SSIM} \\ \midrule
Donut &
  \cellcolor[HTML]{FFBE42}23.43 &
  \cellcolor[HTML]{FFBE42}0.783 &
  \cellcolor[HTML]{FFCA66}22.27 &
  \cellcolor[HTML]{FFCA66}0.693 &
  \cellcolor[HTML]{FFD689}21.39 &
  \cellcolor[HTML]{FFD689}0.583 &
  \cellcolor[HTML]{FEEAC6}20.30 &
  \cellcolor[HTML]{FEEAC6}0.554 &
  \cellcolor[HTML]{F9A400}26.45 &
  \cellcolor[HTML]{F9A400}0.797 &
  \cellcolor[HTML]{FFB21E}25.45 &
  \cellcolor[HTML]{FFB21E}0.773 &
  \cellcolor[HTML]{FFBE42}24.57 &
  \cellcolor[HTML]{FFBE42}0.753 &
  \cellcolor[HTML]{FFD689}23.41 &
  \cellcolor[HTML]{FFD689}0.723 \\
Apple &
  \cellcolor[HTML]{FFBE42}25.28 &
  \cellcolor[HTML]{FFBE42}0.794 &
  \cellcolor[HTML]{FFCA66}24.26 &
  \cellcolor[HTML]{FFCA66}0.703 &
  \cellcolor[HTML]{FFD689}23.30 &
  \cellcolor[HTML]{FFD689}0.682 &
  \cellcolor[HTML]{FEEAC6}22.04 &
  \cellcolor[HTML]{FEEAC6}0.653 &
  \cellcolor[HTML]{F9A400}27.71 &
  \cellcolor[HTML]{F9A400}0.808 &
  \cellcolor[HTML]{FFB21E}26.77 &
  \cellcolor[HTML]{FFB21E}0.789 &
  \cellcolor[HTML]{FFBE42}25.95 &
  \cellcolor[HTML]{FFBE42}0.774 &
  \cellcolor[HTML]{FFD689}24.71 &
  \cellcolor[HTML]{FFD689}0.745 \\
Hydrant &
  \cellcolor[HTML]{FFBE42}22.36 &
  \cellcolor[HTML]{FFBE42}0.850 &
  \cellcolor[HTML]{FFCA66}21.36 &
  \cellcolor[HTML]{FFCA66}0.740 &
  \cellcolor[HTML]{FFD689}20.70 &
  \cellcolor[HTML]{FFD689}0.722 &
  \cellcolor[HTML]{FEEAC6}19.96 &
  \cellcolor[HTML]{FEEAC6}0.692 &
  \cellcolor[HTML]{F9A400}26.03 &
  \cellcolor[HTML]{F9A400}0.862 &
  \cellcolor[HTML]{FFB21E}25.14 &
  \cellcolor[HTML]{FFB21E}0.851 &
  \cellcolor[HTML]{FFBE42}24.47 &
  \cellcolor[HTML]{FFBE42}0.840 &
  \cellcolor[HTML]{FFD689}23.60 &
  \cellcolor[HTML]{FFD689}0.814 \\
Vase &
  \cellcolor[HTML]{FFBE42}24.43 &
  \cellcolor[HTML]{FFBE42}0.826 &
  \cellcolor[HTML]{FFCA66}22.45 &
  \cellcolor[HTML]{FFCA66}0.700 &
  \cellcolor[HTML]{FFD689}20.99 &
  \cellcolor[HTML]{FFD689}0.647 &
  \cellcolor[HTML]{FEEAC6}19.52 &
  \cellcolor[HTML]{FEEAC6}0.586 &
  \cellcolor[HTML]{F9A400}26.94 &
  \cellcolor[HTML]{F9A400}0.847 &
  \cellcolor[HTML]{FFB21E}25.33 &
  \cellcolor[HTML]{FFB21E}0.808 &
  \cellcolor[HTML]{FFBE42}23.71 &
  \cellcolor[HTML]{FFBE42}0.766 &
  \cellcolor[HTML]{FFD689}21.93 &
  \cellcolor[HTML]{FFD689}0.699 \\
Cake &
  \cellcolor[HTML]{FFBE42}22.03 &
  \cellcolor[HTML]{FFBE42}0.801 &
  \cellcolor[HTML]{FFCA66}20.25 &
  \cellcolor[HTML]{FFCA66}0.678 &
  \cellcolor[HTML]{FFD689}18.82 &
  \cellcolor[HTML]{FFD689}0.500 &
  \cellcolor[HTML]{FEEAC6}17.52 &
  \cellcolor[HTML]{FEEAC6}0.461 &
  \cellcolor[HTML]{F9A400}26.10 &
  \cellcolor[HTML]{F9A400}0.775 &
  \cellcolor[HTML]{FFB21E}24.63 &
  \cellcolor[HTML]{FFB21E}0.739 &
  \cellcolor[HTML]{FFBE42}23.26 &
  \cellcolor[HTML]{FFBE42}0.706 &
  \cellcolor[HTML]{FFD689}21.48 &
  \cellcolor[HTML]{FFD689}0.650 \\
Ball &
  \cellcolor[HTML]{FFBE42}23.29 &
  \cellcolor[HTML]{FFBE42}0.742 &
  \cellcolor[HTML]{FFCA66}21.54 &
  \cellcolor[HTML]{FFCA66}0.602 &
  \cellcolor[HTML]{FFD689}20.45 &
  \cellcolor[HTML]{FFD689}0.573 &
  \cellcolor[HTML]{FEEAC6}19.33 &
  \cellcolor[HTML]{FEEAC6}0.538 &
  \cellcolor[HTML]{F9A400}25.35 &
  \cellcolor[HTML]{F9A400}0.767 &
  \cellcolor[HTML]{FFB21E}23.82 &
  \cellcolor[HTML]{FFB21E}0.731 &
  \cellcolor[HTML]{FFBE42}22.71 &
  \cellcolor[HTML]{FFBE42}0.702 &
  \cellcolor[HTML]{FFD689}21.46 &
  \cellcolor[HTML]{FFD689}0.662 \\
Bench &
  \cellcolor[HTML]{FFBE42}20.73 &
  \cellcolor[HTML]{FFBE42}0.794 &
  \cellcolor[HTML]{FFCA66}19.39 &
  \cellcolor[HTML]{FFCA66}0.624 &
  \cellcolor[HTML]{FFD689}18.36 &
  \cellcolor[HTML]{FFD689}0.532 &
  \cellcolor[HTML]{FEEAC6}17.30 &
  \cellcolor[HTML]{FEEAC6}0.483 &
  \cellcolor[HTML]{F9A400}21.96 &
  \cellcolor[HTML]{F9A400}0.708 &
  \cellcolor[HTML]{FFB21E}20.64 &
  \cellcolor[HTML]{FFB21E}0.651 &
  \cellcolor[HTML]{FFBE42}19.56 &
  \cellcolor[HTML]{FFBE42}0.604 &
  \cellcolor[HTML]{FFD689}18.37 &
  \cellcolor[HTML]{FFD689}0.552 \\
Suitcase &
  \cellcolor[HTML]{FFBE42}25.08 &
  \cellcolor[HTML]{FFBE42}0.809 &
  \cellcolor[HTML]{FFCA66}23.34 &
  \cellcolor[HTML]{FFCA66}0.629 &
  \cellcolor[HTML]{FFD689}22.09 &
  \cellcolor[HTML]{FFD689}0.596 &
  \cellcolor[HTML]{FEEAC6}20.74 &
  \cellcolor[HTML]{FEEAC6}0.556 &
  \cellcolor[HTML]{F9A400}28.46 &
  \cellcolor[HTML]{F9A400}0.837 &
  \cellcolor[HTML]{FFB21E}26.74 &
  \cellcolor[HTML]{FFB21E}0.809 &
  \cellcolor[HTML]{FFBE42}25.20 &
  \cellcolor[HTML]{FFBE42}0.779 &
  \cellcolor[HTML]{FFD689}23.71 &
  \cellcolor[HTML]{FFD689}0.739 \\
Teddybear &
  \cellcolor[HTML]{FFBE42}20.87 &
  \cellcolor[HTML]{FFBE42}0.749 &
  \cellcolor[HTML]{FFCA66}19.40 &
  \cellcolor[HTML]{FFCA66}0.568 &
  \cellcolor[HTML]{FFD689}18.28 &
  \cellcolor[HTML]{FFD689}0.533 &
  \cellcolor[HTML]{FEEAC6}17.03 &
  \cellcolor[HTML]{FEEAC6}0.488 &
  \cellcolor[HTML]{F9A400}24.34 &
  \cellcolor[HTML]{F9A400}0.763 &
  \cellcolor[HTML]{FFB21E}22.64 &
  \cellcolor[HTML]{FFB21E}0.714 &
  \cellcolor[HTML]{FFBE42}21.01 &
  \cellcolor[HTML]{FFBE42}0.659 &
  \cellcolor[HTML]{FFD689}19.30 &
  \cellcolor[HTML]{FFD689}0.593 \\
Plant &
  \cellcolor[HTML]{FFBE42}22.60 &
  \cellcolor[HTML]{FFBE42}0.662 &
  \cellcolor[HTML]{FFCA66}21.21 &
  \cellcolor[HTML]{FFCA66}0.565 &
  \cellcolor[HTML]{FFD689}20.16 &
  \cellcolor[HTML]{FFD689}0.531 &
  \cellcolor[HTML]{FEEAC6}19.12 &
  \cellcolor[HTML]{FEEAC6}0.491 &
  \cellcolor[HTML]{F9A400}24.32 &
  \cellcolor[HTML]{F9A400}0.708 &
  \cellcolor[HTML]{FFB21E}22.74 &
  \cellcolor[HTML]{FFB21E}0.663 &
  \cellcolor[HTML]{FFBE42}21.70 &
  \cellcolor[HTML]{FFBE42}0.622 &
  \cellcolor[HTML]{FFD689}20.44 &
  \cellcolor[HTML]{FFD689}0.572 \\ \bottomrule
\end{tabular}
}
\caption{Comparison of PSNR and SSIM Values for Meta-Training with 3 Views and 6 Views. The table presents the Peak Signal-to-Noise Ratio (PSNR) and Structural Similarity Index (SSIM) values obtained through meta-training using 3 views and 6 views, without dependency on external models. Additionally, the table reports average PSNR and SSIM values by selecting the top quartiles from the evaluated scenes, demonstrating the impact of varying view counts on image quality enhancement.}
\label{tab:my-table}
\end{table*}


\end{document}